\documentclass[11pt,a4paper, dvipsnames]{article}
\usepackage[hyperref]{acl2021}
\usepackage{xcolor}
\usepackage{times}
\usepackage{ragged2e}
\usepackage{amsmath}
\usepackage{bm}
\usepackage{latexsym}
\usepackage{subfigure}
\usepackage{graphicx}
\usepackage{float}
\usepackage{longtable}
\usepackage{booktabs} 
\usepackage{multirow}
\usepackage{amssymb}
\usepackage[ruled,vlined]{algorithm2e}
\usepackage{color, soul}

\SetCommentSty{mycommfont}

\usepackage{microtype}

\aclfinalcopy 
\def\aclpaperid{815} 
\definecolor{mygreen}{RGB}{0,220,70}

\title{\textsc{Few-NERD}: A Few-shot Named Entity Recognition Dataset}

\author{Ning Ding$^{1,3}$\thanks{\quad equal contributions}\hspace{0.5em}, Guangwei Xu$^{2*}$,  Yulin Chen$^{3*}$, {Xiaobin Wang}$^{2}$, \\  \textbf{Xu Han}$^{1}$,  \textbf{Pengjun Xie}$^{2}$, \textbf{Hai-Tao Zheng}$^{3\dag}$, \textbf{Zhiyuan Liu}$^{1}$\thanks{\quad corresponding authors}\hspace{0.5em} \\
$^{1}$Department of Computer Science and Technology, Tsinghua University\\ $^{2}$Alibaba Group,
$^{3}$Shenzhen International Graduate School, Tsinghua University \\
\texttt{\{dingn18, yl-chen17, hanxu17\}@mails.tsinghua.edu.cn} \\
\texttt{\{kunka.xgw, xuanjie.wxb, chengchen.xpj\}@alibaba-inc.com} \\
\texttt{\{zheng.haitao\}@sz.tsinghua.edu.cn}, 
\texttt{\{liuzy\}@tsinghua.edu.cn} \\ 
\texttt{\url{https://ningding97.github.io/fewnerd/}}
}

\date{}

\begin{document}
\maketitle
\begin{abstract}


Recently, considerable literature has grown up around the theme of few-shot named entity recognition (NER), but little published benchmark data specifically focused on the practical and challenging task. Current approaches collect existing supervised NER datasets and re-organize them into the few-shot setting for empirical study. These strategies conventionally aim to recognize coarse-grained entity types with few examples, while in practice, most unseen entity types are fine-grained. 
In this paper, we present \textsc{Few-NERD}, a large-scale human-annotated few-shot NER dataset with a hierarchy of 8 coarse-grained and 66 fine-grained entity types.
\textsc{Few-NERD} consists of 188,238 sentences from Wikipedia, 4,601,160 words are included and each is annotated as context or a part of a two-level entity type. 
To the best of our knowledge, this is the first few-shot NER dataset and the largest human-crafted NER dataset. 
We construct benchmark tasks with different emphases to comprehensively assess the generalization capability of models. 
Extensive empirical results and analysis show that \textsc{Few-NERD} is challenging and the problem requires further research. We make \textsc{Few-NERD} public at {\url{https://ningding97.github.io/fewnerd/}}.
 \footnote{The baselines are available at {\url{https://github.com/thunlp/Few-NERD}}}
\end{abstract}

\section{Introduction}

Named entity recognition (NER), as a fundamental task in information extraction, aims to locate and classify named entities from unstructured natural language. 
A considerable number of approaches equipped with deep neural networks have shown promising performance~\cite{chiu2016named} on fully supervised NER. Notably, pre-trained language models (e.g., BERT~\cite{devlin2018bert}) with an additional classifier achieve significant success on this task and gradually become the base paradigm. Such studies demonstrate that deep models could yield remarkable results accompanied by a large amount of annotated corpora.


\begin{figure}[t]
    \centering
    \includegraphics[width = 0.95\linewidth]{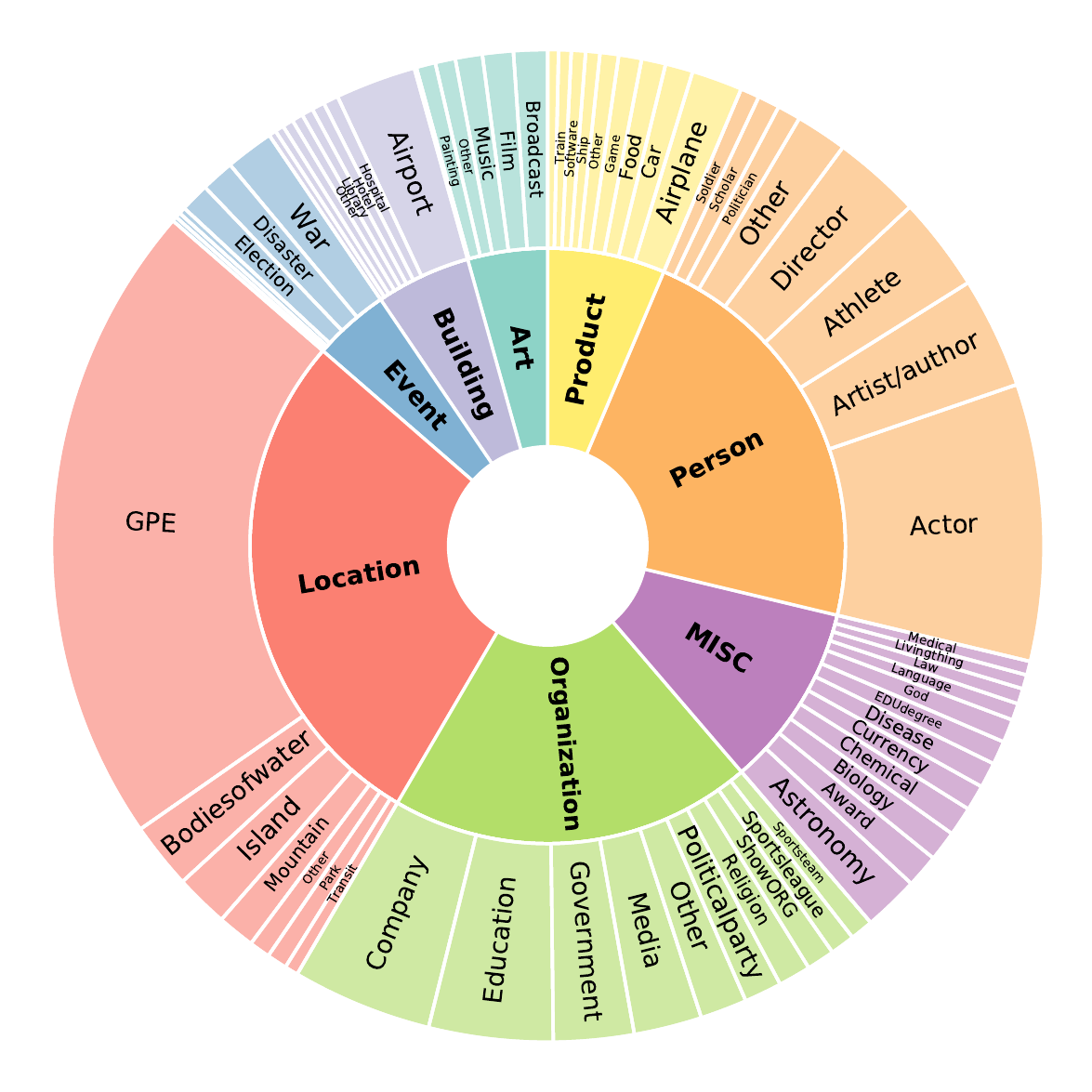}
    \caption{An overview of \textsc{Few-NERD}. The inner circle represents the coarse-grained entity types and the outer circle represents the fine-grained entity types, some types are denoted by abbreviations.}
    \label{fig:overview}
\end{figure}

With the emerging of knowledge from various domains, named entities, especially ones that need professional knowledge to understand, are difficult to be manually annotated on a large scale.
Under this circumstance, studying NER systems that could learn unseen entity types with few examples, i.e., few-shot NER, plays a critical role in this area. There is a growing body of literature that recognizes the importance of few-shot NER and contributes to the task~\cite{hofer2018few, fritzler2019few, yang2020simple, li2020few, huang2020few}. Unfortunately, \textit{there is still no dataset specifically designed for few-shot NER}. Hence, these methods collect previously proposed supervised NER datasets and re-organize them into a few-shot setting. Common options of datasets include OntoNotes~\cite{weischedel2013ontonotes}, CoNLL'03~\cite{sang2003introduction}, WNUT'17~\cite{derczynski2017results}, etc. 
These research efforts of few-shot learning for named entities mainly face two challenges:
First, most datasets used for few-shot learning have only 4-18 coarse-grained entity types, making it hard to construct an adequate variety of ``N-way'' meta-tasks and learn correlation features. And in reality, we observe that most unseen entities are fine-grained. Second, because of the lack of benchmark datasets, the settings of different works are inconsistent~\cite{huang2020few,yang2020simple}, leading to unclear comparisons. 
To sum up, these methods make promising contributions to few-shot NER, nevertheless, a specific dataset is urgently needed to provide a unified benchmark dataset for rigorous comparisons. 

To alleviate the above challenges, we present a large-scale human-annotated few-shot NER dataset, $\textsc{Few-NERD}$, which consists of 188.2k sentences extracted from the Wikipedia articles and 491.7k entities are manually annotated by well-trained annotators (Section~\ref{sec:human}). To the best of our knowledge, $\textsc{Few-NERD}$ is the first dataset specially constructed for few-shot NER and also one of the largest human-annotated NER dataset (statistics in Section~\ref{sec:datasize}). We carefully design an annotation schema of 8 coarse-grained entity types and 66 fine-grained entity types by conducting several pre-annotation rounds. (Section~\ref{sec:schema}). In contrast, as the most widely-used NER datasets, CoNLL has 4 entity types, WNUT'17 has 6 entity types and OntoNotes has 18 entity types (7 of them are value types). The variety of entity types makes $\textsc{Few-NERD}$ contain rich contextual features with a finer granularity for better evaluation of few-shot NER. 
The distribution of the entity types in $\textsc{Few-NERD}$ is shown in Figure~\ref{fig:overview}, more details are reported in Section~\ref{sec:dist}. We conduct an analysis of the mutual similarities among all the entity types of $\textsc{Few-NERD}$ to study knowledge transfer (Section~\ref{sec:sim}). The results show that our dataset can provide sufficient correlation information between different entity types for few-shot learning.


For benchmark settings, we design three tasks on the basis of $\textsc{Few-NERD}$, including a standard supervised task (\textsc{Few-NERD  (sup)}) and two few-shot tasks ($\textsc{Few-NERD-intra}$) and \textsc{Few-NRTD (inter)}), for more details see Section~\ref{sec:bench}. \textsc{Few-NERD (sup)}, \textsc{Few-NERD (intra)}, and \textsc{Few-NERD (inter)} assess instance-level generalization, type-level generalization and knowledge transfer of NER methods, respectively. We implement models based on the recent state-of-the-art approaches and evaluate them on $\textsc{Few-NERD}$ (Section~\ref{sec:exp}). And empirical results show that $\textsc{Few-NERD}$ is challenging on all these three settings. We also conduct sets of subsidiary experiments to analyze promising directions of few-shot NER. Hopefully, the research of few-shot NER could be further facilitated by $\textsc{Few-NERD}$.

\section{Related Work}

As a pivotal task of information extraction, NER is essential for a wide range of technologies~\cite{cui2019kbqa, li2019chinese, ding-etal-2019-event, shen2020modeling}. And a considerable number of NER datasets have been proposed over the years. For example, CoNLL'03~\cite{sang2003introduction} is regarded as one of the most popular datasets, which is curated from Reuters News and includes 4 coarse-grained entity types. Subsequently, a series of NER datasets from various domains are proposed~\cite{balasuriya2009named, ritter-etal-2011-named, weischedel2013ontonotes, stubbs2015annotating, derczynski2017results}. These datasets formulate a sequence labeling task and most of them contain 4-18 entity types. Among them, due to the high quality and size, OntoNotes 5.0~\cite{weischedel2013ontonotes} is considered as one of the most widely used NER datasets recently.

As approaches equipped with deep neural networks have shown satisfactory performance on NER with sufficient supervision~\cite{lample-etal-2016-neural, ma2016end}, few-shot NER has received increasing attention~\cite{hofer2018few, fritzler2019few, yang2020simple, li2020few}. Few-shot NER is a considerably challenging and practical problem that could facilitate the understanding of textual knowledge for neural model~\cite{huang2020few}. Due to the lack of specific benchmarks of few-shot NER, current methods collect existing NER datasets and use different few-shot settings. 
To provide a benchmark that could comprehensively assess the generalization of models under few examples, we annotate \textsc{Few-NERD}. To make the dataset practical and close to reality, we adopt a fine-grained schema of entity annotation, which is inspired and modified from previous fine-grained entity recognition studies~\cite{ling2012fine, gillick2014context, choi2018ultra, ringland2019nne}.

\section{Problem Formulation}
\subsection{Named Entity Recognition}

NER is normally formulated as a sequence labeling problem. Specifically, for an input sequence of tokens $\bm{x}=\{x_1,x_2,...,x_t\}$, NER aims to assign each token $x_i$ a label $y_i \in \mathcal{Y}$ to indicate either the token is a part of a named entity (such as \texttt{Person}, \texttt{Organization}, \texttt{Location}) or not belong to any entities (denoted as \texttt{O} class), $\mathcal{Y}$ being a set of pre-defined entity-types.

\subsection{Few-shot Named Entity Recognition}

$N$-way $K$-shot learning is conducted by iteratively constructing episodes. For each episode in training, $N$ classes ($N$-way) and $K$ examples ($K$-shot) for each class are sampled to build a support set $\mathcal{S}_{\text{train}} = \{\bm{x}^{(i)}, \bm{y}^{(i)}\}_{i=1}^{N*K}$,  and $K'$ examples for each of $N$ classes are sampled to construct a query set $\mathcal{Q}_{\text{train}} = \{\bm{x}^{(j)}, \bm{y}^{(j)}\}_{j=1}^{N*K'}$, and $\mathcal{S} \bigcap \mathcal{Q} = \varnothing$. Few-shot learning systems are trained by predicting labels of query set $\mathcal{Q}_{\text{train}}$ with the information of support set $\mathcal{S}_{\text{train}}$. The supervision of $\mathcal{S}_{\text{train}}$ and $\mathcal{Q}_{\text{train}}$ are available in training. In the testing procedure, all the classes are unseen in the training phase, and by using few labeled examples of support set $\mathcal{S}_{\text{test}}$, few-shot learning systems need to make predictions of the unlabeled query set $\mathcal{Q}_{\text{test}}$ ($\mathcal{S} \bigcap \mathcal{Q} = \varnothing$). However, in the sequence labeling problem like NER, a sentence may contain multiple entities from different classes. And it is imperative to sample examples in sentence-level since contextual information is crucial for sequence labeling problems, especially for NER. Thus the sampling is more difficult than conventional classification tasks like relation extraction~\cite{han2018fewrel}.

Some previous works~\cite{yang2020simple, li2020few} use greedy-based sampling strategies to iteratively judge if a sentence could be added into the support set, but the limitation becomes gradually strict during the sampling. For example, when it comes to a 5-way 5-shot setting, if the support set already had 4 classes with 5 examples and 1 class with 4 examples, the next sampled sentence must only contain the specific one entity to strictly meet the requirement of $5$ way $5$ shot. It is not suitable for $\textsc{Few-NERD}$ since it is annotated with dense entities. Thus, as shown in Algorithm~\ref{alg:gre} we adopt a $N$-way $K$$\sim$$2K$-shot setting in our paper, the primary principle of which is to ensure that each class in $\mathcal{S}$ contain $K$$\sim$$2K$ examples, effectively alleviating the limitations of sampling.

\begin{algorithm}[h]
\caption{Greedy $N$-way $K$$\sim$$2K$-shot sampling algorithm}
\LinesNumbered 
\label{alg:gre}
\KwIn{Dataset $\mathcal{X}$, Label set $\mathcal{Y}$, $N$, $K$}
\KwOut{output result}
$\mathcal{S}\leftarrow \varnothing$;  {\tcp{Init the support set}} 
{\tcp{Init the count of entity types}} 
\For{$i=1$ to $N$}{
  $\text{Count}[i] = 0$ \;
}
\Repeat{$\text{Count}_i \geq K$ for $i = 1$ to N }{
Randomly sample $(\bm{x}, \bm{y}) \in \mathcal{X}$ \; 

Compute $|\text{Count}|$ and $\text{Count}_i$ after update \;
    \eIf{$|\text{Count}| > N$ or $\exists \text{Count}[i] > 2K$}{
      Continue \;
    }{
      $\mathcal{S} = \mathcal{S} \bigcup (\bm{x}, \bm{y})$ \;
      Update $\text{Count}_i$ \;
    }
}
\end{algorithm}
\vspace{-0.2cm}

\section{Collection of \textsc{Few-NERD}}

\subsection{Schema of Entity Types}
\label{sec:schema}

The primary goal of \textsc{Few-NERD} is to construct a fine-grained dataset that could specifically be used in the few-shot NER scenario. 
Hence, schemas of traditional NER datasets such as CoNLL'03, OntoNotes that only contain 4-18 coarse-grained types could not meet the requirements. 
The schema of \textsc{Few-NERD} is inspired by \textsc{Figer}~\cite{ling2012fine}, which contains 112 entity tags with good coverage. On this basis, we make some modifications according to the practical situation. It is worth noting that $\textsc{Few-NERD}$ focuses on named entities, omitting value/numerical/time/date entity types~\cite{weischedel2013ontonotes, ringland2019nne} like \texttt{Cardinal, Day, Percent}, etc.

First, we modify the \textsc{Figer} schema into a two-level hierarchy to incorporate simple domain information~\cite{gillick2014context}. The coarse-grained types are \{\texttt{Person}, \texttt{Location}, \texttt{Organization}, \texttt{Art}, \texttt{Building}, \texttt{Product}, \texttt{Event}, \texttt{Miscellaneous} \}.
Then we statistically count the frequency of entity types in the automatically annotated \textsc{Figer}. By removing entity types with low frequency, there are 80 fine-grained types remaining. Finally, to ensure the practicality of the annotation process, we conduct rounds of pre-annotation and make further modifications to the schema. For example, we combine the types of \texttt{Country}, \texttt{Province/State}, \texttt{City}, \texttt{Restrict} into a class \texttt{GPE}, since it is difficult to distinguish these types only based on context (especially GPEs at different times). For another example, we create a \texttt{Person-Scholar} type, because in the pre-annotation step, we found that there are numerous person entities that express the semantics of research, such as mathematician, physicist, chemist, biologist, paleontologist, but the Figer schema does not define this kind of entity type. We also conduct rounds of manual denoising to select types with truly high frequency.

Consequently, the finalized schema of \textsc{Few-NERD} includes 8 coarse-grained types and 66 fine-grained types, which is detailedly shown accompanied by selected examples in Appendix.


\subsection{Paragraph Selection}
\label{sec:para}

The raw corpus we use is the entire Wikipedia dump in English, which has been widely used in constructions of NLP datasets~\cite{han2018fewrel,yang2018hotpotqa, wang2020maven}. Wikipedia contains a large variety of entities and rich contextual information for each entity.

\textsc{Few-NERD} is annotated in paragraph-level, and it is crucial to effectively select paragraphs with sufficient entity information. Moreover, the category distribution of the data is expected to be balanced since the data is applied in a few-shot scenario. It is also a key difference between $\textsc{Few-NERD}$ and previous NER datasets, whose entity distributions are usually considerably uneven. In order to do so, we construct a dictionary for each fine-grained type by automatically collecting entity mentions annotated in $\textsc{Figer}$, then the dictionaries are manually denoised.
We develop a search engine to retrieve paragraphs including entity mentions of the distant dictionary. For each entity, we choose 10 paragraphs and construct a candidate set. Then, for each fine-grained class, we randomly select 1000 paragraphs for manual annotation. Eventually, 66,000 paragraphs are selected, consisting of 66 fine-grained entity types, and each paragraph contains an average of 61.3 tokens.

\subsection{Human Annotation}
\label{sec:human}

As named entities are expected to be context-dependent, annotation of named entities is complicated, especially with such a large number of entity types. For example, shown in Table~\ref{tab:case}, ``\textit{London is the fifth album by the British rock band Jesus Jones..}'', where \textit{London} should be annotated as an entity of \texttt{Art-Music} rather than \texttt{Location-GPE}. Such a situation requires that the annotator has basic linguistic training and can make reasonable judgments based on the context.

\begin{table}[]
    \centering
    \scalebox{0.9}{
    \begin{tabular}{p{7.7cm}} \toprule
        \textbf{Paragraph} \\\midrule
        \ \ ${\color{Plum}\textit{London}_{\texttt{[Art-Music]}}}$ is the fifth album by the  ${\color{BlueViolet}\textit{British}_{\texttt{[Loc-GPE]}}}$  rock band ${\color{PineGreen}\textit{Jesus Jones}_{\texttt{[Org-ShowOrg]}}}$ in 2001 through ${\color{Mahogany}\textit{Koch Records}_{\texttt{[Org-Company]}}}$. Following the commercial failure of 1997's "${\color{Plum}\textit{Already}_{\texttt{[Art-Music]}}}$" which led to the band and ${\color{Mahogany}\textit{EMI}_{\texttt{[Org-Company]}}}$ parting ways,
        the band took a hiatus before regathering for the recording of "${\color{Plum}\textit{London}_{\texttt{[Art-Music]}}}$" for Koch/Mi5 Recordings, with a more alternative rock approach as opposed to the techno sounds on their previous albums. The album had low-key promotion, initially only being released in the ${\color{BlueViolet}\textit{United States}_{\texttt{[Loc-GPE]}}}$. Two EP's were released from the album, "${\color{Plum}\textit{Nowhere Slow}_{\texttt{[Art-Music]}}}$" and "${\color{Plum}\textit{In the Face Of All This}_{\texttt{[Art-Music]}}}$". 
        \\ \bottomrule
    \end{tabular}}
    \caption{An annotated case of $\textsc{Few-NERD}$}
    \label{tab:case}
    \vspace{-0.24cm}
\end{table}

Annotators of $\textsc{Few-NERD}$ include 70 annotators and 10 experienced experts. All the annotators have linguistic knowledge and are instructed with detailed and formal annotation principles. Each paragraph is independently annotated by two well-trained annotators. Then, an experienced expert goes over the paragraph for possible wrong or omissive annotations, and make the final decision. With 70 annotators participated, each annotator spends an average of 32 hours during the annotation process. We ensure that all the annotators are fairly compensated by market price according to their workload (the number of examples per hour).  
The data is annotated and submitted in batches, and each batch contains 1000$\sim$3000 sentences.
To ensure the quality of $\textsc{Few-NERD}$, for each batch of data, we randomly select 10\% sentences and conduct double-checking. If the accuracy of the annotation is lower than 95 \% (measured in sentence-level), the batch will be re-annotated. Furthermore, we calculate the Cohen's Kappa~\cite{cohen1960coefficient} to measure the aggreements between two annotators, the result is 76.44\%, which indicates a high degree of consistency.


\section{Data Analysis}


\begin{table*}[]
\centering
\scalebox{0.86}{
\begin{tabular}{lrrrrr}
\toprule \textbf{Datasets}      & \# \textbf{Sentences} & \# \textbf{Tokens} & \# \textbf{Entities} & \# \textbf{Entity Types} & \textbf{Domain}       \\ \midrule
CoNLL'03~\cite{sang2003introduction}   & 22.1k             & 301.4k          & 35.1k                & 4            & Newswire     \\
WikiGold~\cite{balasuriya2009named}      & 1.7k             & 39k          & 3.6k                & 4            & General      \\
OntoNotes~\cite{weischedel2013ontonotes}& 103.8k             & 2067k          & 161.8k                & 18            & General      \\
WNUT'17~\cite{derczynski2017results}     & 4.7k             & 86.1k          & 3.1k                & 6            & SocialMedia \\
I2B2~\cite{stubbs2015annotating}          & 107.9k             & 805.1k          & 28.9k                & 23            & Medical      \\ \midrule
\textsc{Few-NERD}      &     \textbf{188.2k}         & \textbf{4601.2k}         &        \textbf{491.7k}          &    \textbf{66}         & General \\ \bottomrule     
\end{tabular}}
\caption{Statistics of $\textsc{Few-NERD}$ and multiple widely used NER datasets. For CoNLL'03, WikiGold, and I2B2, we report the statistics in the original paper. For OntoNotes 5.0 (LDC2013T19), we download and count all the data (English) annotated by the NER labels, some works use different split of OntoNotes 5.0 and may report different statistics. For WNUT'17, we download and count all the data.}
\label{table:stat}
\vspace{-0.2cm}
\end{table*}

\subsection{Size and Distribution of $\textsc{Few-NERD}$}
\label{sec:datasize}
\label{sec:dist}
$\textsc{Few-NERD}$ is not only the first few-shot dataset for NER, but it also is one of the biggest human-annotated NER datasets. We report the the statistics of the number of sentences, tokens, entity types and entities of \textsc{Few-NERD} and several widely-used NER datasets in Table~\ref{table:stat}, including CoNLL'03, WikiGold, OntoNotes 5.0, WNUT'17 and I2B2. We observe that although OntoNotes and I2B2 are considered as large-scale datasets, $\textsc{Few-NERD}$ is significantly larger than all these datasets. Moreover, $\textsc{Few-NERD}$ contains more entity types and annotated entities.
As introduced in Section~\ref{sec:para}, \textsc{Few-NERD} is designed for few-shot learning and the distribution could not be severely uneven. Hence, we balance the dataset by selecting paragraphs through a distant dictionary. 
The data distribution is illustrated in Figure~\ref{fig:overview}, where $\texttt{Location}$ (especially \texttt{GPE}) and \texttt{Person} are entity types with the most examples.
Although utilizing a distant dictionary to balance the entity types could not produce a fully balanced data distribution, it still ensures that each fine-grained type has a sufficient number of examples for few-shot learning.



\subsection{Knowledge Correlations among Types}
\label{sec:sim}

\begin{figure}
    \centering
    \includegraphics[width = 0.96\linewidth]{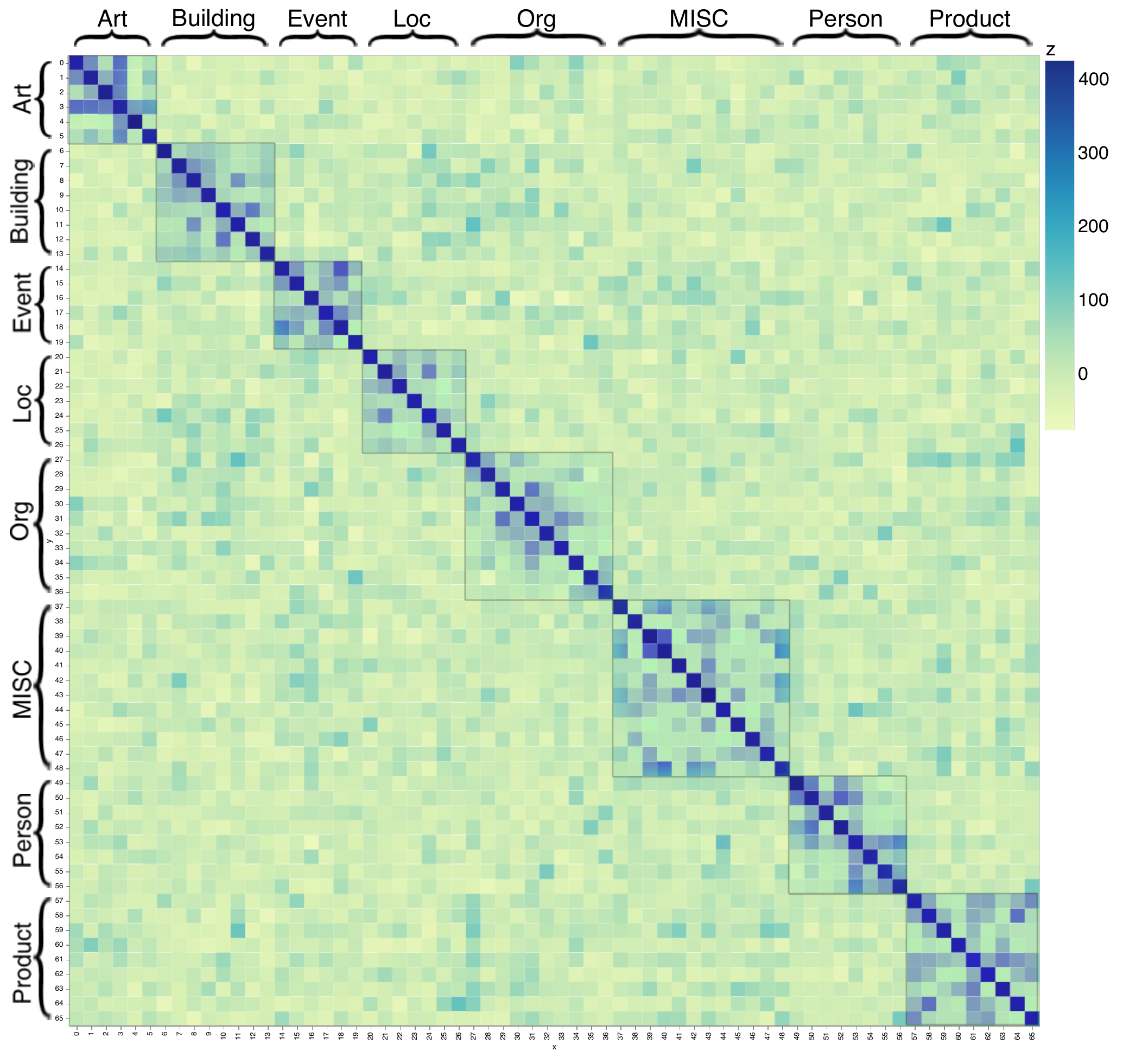}
    \caption{A heat map to illustrate knowledge correlations among type in $\textsc{Few-NERD}$, each small colored square represents the similarity of two entity types.}
    \label{fig:heat}
    \vspace{-0.3cm}
\end{figure}

Knowledge transfer is crucial for few-shot learning~\cite{li2019large}. To explore the knowledge correlations among all the entity types of $\textsc{Few-NERD}$, we conduct an empirical study about entity type similarities in this section.
We train a BERT-Tagger (details in Section~\ref{sec:models}) of 70\% arbitrarily selected data on $\textsc{Few-NERD}$ and use 10\% data to select the model with best performance (it is actually the setting of $\textsc{Few-NERD (sup)}$ in Section~\ref{sec:bench-supner}). After obtaining a contextualized encoder, we produce entity mention representations of the remaining 20\% data of \textsc{Few-NERD}. Then, for each fine-grained types, we randomly select 100 instances of entity embeddings. We mutually compute the dot product among entity embeddings for each type two by two and average them to obtain the similarities among types, which is illustrated in Figure~\ref{fig:heat}. We observe that entity types shared identical coarse-grained types typically have larger similarities, resulting in an easier knowledge transfer. In contrast, although some of the fine-grained types have large similarities, most of them across coarse-grained types share little correlations due to distinct contextual features. This result is consistent with intuition. Moreover, it inspires our benchmark-setting from the perspective of knowledge transfer (see Section~\ref{sec:bench-fewner}).

\section{Benchmark Settings}
\label{sec:bench}


We collect and manually annotate 188,238 sentences with 66 fine-grained entity types in total, which makes $\textsc{Few-NERD}$ one of the largest human-annotated NER datasets. 
To comprehensively exploit such rich information of entities and contexts, as well as evaluate the generalization of models from different perspectives, we construct three tasks based on $\textsc{Few-NERD}$ (Statistics are reported in Table~\ref{tab:stats_bench}).

\subsection{Standard Supervised NER}
\label{sec:bench-supner}
\textbf{\textsc{Few-NERD (sup)}}\quad We first adopt a standard \textit{supervised setting} for NER by randomly splitting 70\% data as the training data, 10\% as the validation data and 20\% as the testing data.
 In this setting, the training set, dev set, and test set contain the whole 66 entity types. Although the supervised setting is not the ultimate goal of the construction of \textsc{Few-NERD}, it is still meaningful to assess the instance-level generalization for NER models. As shown in Section 6.2, due to the large number of entity types, \textsc{Few-NERD} is very challenging even in a standard supervised setting.

\subsection{Few-shot NER}
\label{sec:bench-fewner}
The core intuition of few-shot learning is to learn new classes from few examples. Hence, we first split the overall entity set (denoted as $\mathcal{E}$) into three mutually disjoint subsets, respectively denoted as $\mathcal{E}_{\text{train}}, \mathcal{E}_{\text{dev}}, \mathcal{E}_{\text{test}}$, and $\mathcal{E}_{\text{train}} \bigcup \mathcal{E}_{\text{dev}} \bigcup \mathcal{E}_{\text{test}} = \mathcal{E}$, $\mathcal{E}_\text{train} \bigcap \mathcal{E}_{\text{dev}} \bigcap \mathcal{E}_{\text{test}} = \varnothing$. Note that all the entity types are fine-grained types.
Under this circumstance, instances in train, dev and test datasets only consist of instances with entities in $\mathcal{E}_{\text{train}}, \mathcal{E}_{\text{dev}}, \mathcal{E}_{\text{test}}$ respectively. However, NER is a sequence labeling problem, and it is possible that a sentence contains several different entities. To avoid the observation of new entity types in the training phase, we replace the labels of entities that belong to $\mathcal{E}_{\text{test}}$ with \texttt{O} in the training set. Similarly, in the test set, entities that belongs to $\mathcal{E}_{\text{train}}$ and $\mathcal{E}_{\text{dev}}$ are also replaced by \texttt{O}. Based on this setting, we develop two few-shot NER tasks adopting different splitting strategies.

\noindent\textbf{\textsc{Few-NERD (intra)}}\quad Firstly, we construct $\mathcal{E}_{\text{train}}$, $\mathcal{E}_{\text{dev}}$ and $\mathcal{E}_{\text{test}}$ according to the coarse-grained types. In other words, all the entities in different sets belong to different coarse-grained types. In the basis of the principle that we should replace as few as possible entities with \texttt{O}, we assign all the fine-grained entity types belonging to \texttt{People, MISC, Art, Product} to $\mathcal{E}_{\text{train}}$, all the fine-grained entity types belonging to \texttt{Event, Building} to  $\mathcal{E}_{\text{dev}}$, and all the fine-grained entity types belonging to \texttt{ORG, LOC} to  $\mathcal{E}_{\text{test}}$, respectively. Based on Figure~\ref{fig:heat}, in this setting, the training set, dev set and test set share little knowledge, making it a difficult benchmark.

\noindent\textbf{\textsc{Few-NERD (inter)}}\quad In this task, although all the fine-grained entity types are mutually disjoint in $\mathcal{E}_{\text{train}}$, $\mathcal{E}_{\text{dev}}$, the coarse-grained types are shared. Specifically, we roughly assign 60\% fine-grained types of all the 8 coarse-grained types to $\mathcal{E}_{\text{train}}$, 20\% to $\mathcal{E}_{\text{dev}}$ and 20\% $\mathcal{E}_{\text{test}}$, respectively. The intuition of this setting is to explore if the coarse information will affect the prediction of new entities.



\begin{table}[]
    \centering
    \scalebox{0.88}{
    \begin{tabular}{lccc}
        \toprule
        \textbf{Split} & \textbf{\#Train} & \textbf{\#Dev} & \textbf{\#Test} \\
        \midrule
        \textsc{Few-NERD (sup)} & 131,767 & 18,824 & 37,648\\
        \textsc{Few-NERD (intra)} & 99,519 & 19,358 & 44,059\\
        \textsc{Few-NERD (inter)} & 130,112 & 18,817 & 14,007\\ \bottomrule
    \end{tabular}}
    \caption{Statistics of train, dev and test sets for three tasks of $\textsc{Few-NERD}.$ We remove the sentences with no entities for the few-shot benchmarks.}
    \label{tab:stats_bench}
    \vspace{-0.3cm}
\end{table}

\section{Experiments}

\label{sec:exp}

\subsection{Models}
\label{sec:models}

Recent studies show that pre-trained language models with deep transformers (e.g., BERT~\cite{devlin2018bert}) have become a strong encoder for NER~\cite{li2020unified}. We thus follow the empirical settings and use BERT as the backbone encoder in our experiments. We denote the parameters as $\theta$ and the encoder as $f_\theta$. Given a sequence $\bm{x} = \{x_1,...,x_n\}$,
for each token $x_i$, the encoder produces contextualized representations as:
\begin{equation}
	\bm{h} = [\bm{h}_1,...,\bm{h}_n] = f_\theta([x_1,...,x_n]).
\end{equation}
Specifically, we implement four BERT-based models for supervised and few-shot NER, which are BERT-Tagger~\cite{devlin-etal-2019-bert}, ProtoBERT~\cite{snell2017prototypical}, NNShot~\cite{yang2020simple} and StructShot~\cite{yang2020simple}.

\noindent \textbf{BERT-Tagger} \quad As stated in Section~\ref{sec:bench-supner}, we construct a standard supervised task based on $\textsc{Few-NERD}$, thus we implement a simple but strong baseline BERT-Tagger for supervised NER. BERT-Tagger is built by adding a linear classifier on top of BERT and trained with a cross-entropy objective under a full supervision setting. 

\noindent \textbf{ProtoBERT}\quad Inspired by achievements of meta-learning approaches~\cite{finn2017model, snell2017prototypical,ding2021prototypical} on few-shot learning.
The first baseline model we implement is ProtoBERT, which is a method based on prototypical network~\cite{snell2017prototypical} with a backbone of BERT~\cite{devlin2018bert} encoder. This approach derives a prototype $\bm{z}$ for each entity type by computing the average of the embeddings of the tokens that share the same entity type. The computation is conducted in support set $\mathcal{S}$. For the $i$-th type, the prototype is denoted as $\bm{z}_i$ and the support set is $\mathcal{S}_i$,
\begin{equation}
	\bm{z}_i = \frac{1}{|\mathcal{S}_i|} \sum_{{x} \in \mathcal{S}_i} f_\theta(x).
\end{equation}
While in the query set $\mathcal{Q}$, for each token $x \in \mathcal{Q}$, we firstly compute the distance between $x$ and all the prototypes. We use the $l$-2 distance as the metric function $d(f_\theta(x), \bm{z}) = ||f_\theta(x) - \bm{z}||^2_2$. Then, through the distances between $x$ and all other prototypes, we compute the prediction probability of $x$ over all types. In the training step, parameters are updated in each meta-task. In the testing step, the prediction is the label of the nearest prototype to $x$. That is, for a support set $\mathcal{S}_\mathcal{Y}$ with types of $\mathcal{Y}$ and a query $x$, the prediction process is given as
\begin{equation}
\begin{split}
    y^* &= \text{arg} \min_{y \in \mathcal{Y}} d_y(x), \\
    d_y(x) &= d(f_\theta(x), \bm{z}_y).
\end{split}
\end{equation}

\noindent \textbf{NNShot \& StructShot} \quad NNShot and StructShot~\cite{yang2020simple} are the state-of-the-art methods based on token-level nearest neighbor classification. In our experiments, we use BERT as the backbone encoder to produce contextualized representations for fair comparison. Different from the prototype-based method, NNShot determines the tag of one query based on the token-level distance, which is computed as $d(f_\theta(x), f_\theta(x')) = ||f_\theta(x) - f_\theta(x')||^2_2$. Hence, for a support set $\mathcal{S}_\mathcal{Y}$ with type of $\mathcal{Y}$ and a query $x$,
\begin{equation}
    \begin{split}
    y^* &= \text{arg} \min_{y \in \mathcal{Y}} d_y(x), \\
    d_y(x) &= \min_{x' \in \mathcal{S}_y} d(f_\theta(x), f_\theta(x')).
\end{split}
\end{equation}

With the identical basic structure as NNShot, StructShot adopts an additional Viterbi decoder during the inference phase~\cite{hou2020few} (not in training phase), where we estimate a transition distribution $p(y'|y)$ and an emission distribution $p(y|x)$ and  solve the problem:
\begin{equation}
    \bm{y}^* = \text{arg} \max_{\bm{y}} \prod^T_{t=1} p(y_t|x) \times p(y_t|y_{t-1}).
\end{equation}

To sum up, BERT-Tagger is a well-acknowledged baseline that could produce pronounced results on supervised NER. ProtoBERT, and NNShot \& StructShot respectively use prototype-level and token-level similarity scores to tackle the few-shot NER problem. These baselines are strong and representative models of the NER task. For implementation details, please refer to Appendix.

We evaluate models by considering query sets $\mathcal{Q}_{\text{test}}$ of test episodes.
We calculate the precision (P), recall (R) and micro F1-score over all test episodes.
Instead of the popular BIO schema, we utilize the IO schema in our experiments, using $\texttt{I-type}$ to denote all the tokens of a named entity and $\texttt{O}$ to denote other tokens.


\begin{figure*}[ht]
    \centering
    \includegraphics[width = 1\linewidth]{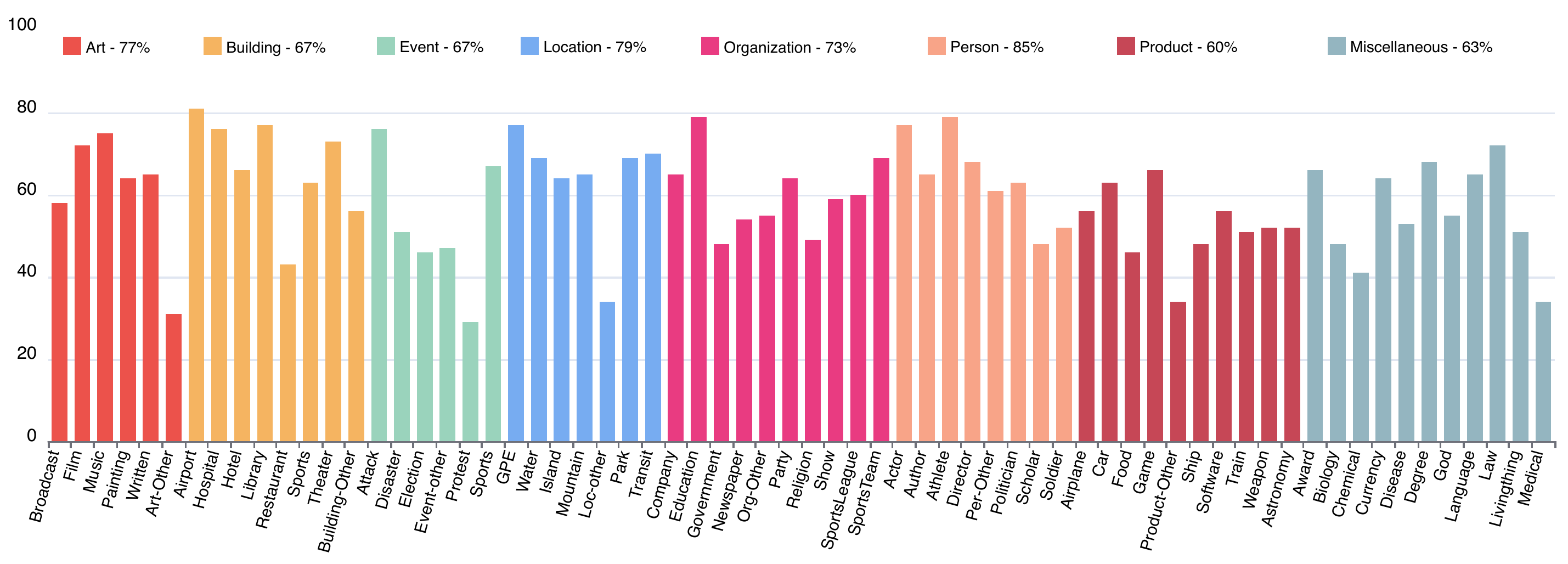}
    \caption{F1-scores of different entity types on \textsc{Few-NERD (SUP)}, we report the average performance of each coarse-grained entity type on the legends.}
    \label{fig:types}
\end{figure*}

\begin{table}[]
    \centering
    \scalebox{0.86}{
    \begin{tabular}{lccc}
        \toprule
        \textbf{Datasets} & \textbf{P} & \textbf{R} & \textbf{F1} \\
        \cmidrule(r){1-1} \cmidrule(r){2-4}
        CoNLL'03 &90.62& 92.07&  91.34\\
        OntoNotes 5.0 & 90.00 & 88.24 & 89.11\\
        \cmidrule(r){1-1} \cmidrule(r){2-4}
        \textsc{{Few-NERD (sup)}} & 65.56 ($\color{red}{\downarrow}$) & 68.78 ($\color{red}{\downarrow}$) & 67.13 ($\color{red}{\downarrow}$) \\ \bottomrule
        
    \end{tabular}}
    \caption{Results of BERT-Tagger on previous NER datasets and the supervised setting of $\textsc{Few-NERD}$.}
    \vspace{-0.3cm}
    \label{tab:sup}
\end{table}

\begin{table*}[h]
\centering
\scalebox{0.57}{
\begin{tabular}{lcccccccccccc} \toprule
\multirow{3}{*}{\textbf{\large{Model}}} & \multicolumn{12}{c}{\textbf{\large{\textsc{Few-NERD(intra)}}}}                                               \\ \cmidrule(r){2-4} \cmidrule(r){5-7} \cmidrule(r){8-10} \cmidrule(r){11-13}
                        & \multicolumn{3}{c}{\large{\textbf{5 way 1$\sim$2 shot}}} & \multicolumn{3}{c}{\large{\textbf{5 way 5$\sim$10 shot}}} & \multicolumn{3}{c}{\large{\textbf{10 way 1$\sim$2 shot}}}& \multicolumn{3}{c}{\large{\textbf{10 way 5$\sim$10 shot}}} \\ \cmidrule(r){2-4} \cmidrule(r){5-7} \cmidrule(r){8-10} \cmidrule(r){11-13}
                          & \large{P}            & \large{R}           & \large{F1}           & {P}             & \large{R}            & \large{F1}           & {P}             & \large{R}            & \large{F1}           & \large{P}             & \large{R}            & \large{F1}            \\  \cmidrule(r){1-1} \cmidrule(r){2-4} \cmidrule(r){5-7} \cmidrule(r){8-10} \cmidrule(r){11-13}
Proto     &     15.97±0.61
         &     29.66±1.39        & 20.76±0.84             &    36.34±1.33  &  \textbf{51.32±0.45}            &      \textbf{42.54±0.94}        &       11.33±0.57       &          \textbf{22.47±0.49}    &        15.05±0.44     &     29.39±0.27      &      \textbf{44.51±1.00}        &        \textbf{35.40±0.13}       \\
NNShot                  &      24.15±0.35     &     27.65±1.63   &      25.78±0.91       &     32.91±0.62       &    40.19±1.22       &      36.18±0.79    &     16.25±0.22       &       20.90±1.38      &   18.27±0.41       &     24.86±0.30         &   30.49±0.96     &  27.38±0.53   \\
Struct              &     \textbf{32.99±0.76}         &      \textbf{27.85±0.98}     &      \textbf{30.21±0.90}       &       \textbf{46.78±1.00}       &      32.06±2.17    &       38.00±1.29     &       \textbf{26.05±0.53}       &      17.65±1.34     &    \textbf{21.03±1.13}    &       \textbf{40.88±0.83}      &   19.52±0.49   &     26.42±0.60    \\ \bottomrule
\end{tabular}}
\caption{Performance of state-of-art models on \textsc{Few-NERD (intra)}.}
\label{tab:intra}
\end{table*}

\begin{table*}[!h]
\centering
\scalebox{0.57}{
\begin{tabular}{lcccccccccccc} \toprule
\multirow{3}{*}{\textbf{\large{Model}}} & \multicolumn{12}{c}{\textbf{\textsc{\large{Few-NERD(inter)}}}}                                               \\ \cmidrule(r){2-4} \cmidrule(r){5-7} \cmidrule(r){8-10} \cmidrule(r){11-13}
                        & \multicolumn{3}{c}{\large{\textbf{5 way 1$\sim$2 shot}}} & \multicolumn{3}{c}{\large{\textbf{5 way 5$\sim$10 shot}}} & \multicolumn{3}{c}{\large{\textbf{10 way 1$\sim$2 shot}}}& \multicolumn{3}{c}{\large{\textbf{10 way 5$\sim$10 shot}}} \\ \cmidrule(r){2-4} \cmidrule(r){5-7} \cmidrule(r){8-10} \cmidrule(r){11-13}
                          & \large{P}            & \large{R}           & \large{F1}           & \large{P}             & \large{R}            & \large{F1}           & \large{P}             & \large{R}            & \large{F1}           & \large{P}             & \large{R}            & \large{F1}            \\  \cmidrule(r){1-1} \cmidrule(r){2-4} \cmidrule(r){5-7} \cmidrule(r){8-10} \cmidrule(r){11-13}
Proto     &     32.04±1.75         &     {49.30±0.68}       & 38.83±1.49         &    {52.54±1.32}         &  \textbf{66.76±1.01}          &      \textbf{58.79±0.44}        &      26.02±1.32      &          {43.17±0.92}    &       32.45±0.79    &     46.38±0.42         &      \textbf{61.60±0.36}        &        \textbf{52.92±0.37}       \\
NNShot                  &     42.57±1.27     &     \textbf{53.09±0.54}    &    47.24±1.00       &  51.03±0.63        &   61.15±0.63        &    55.64±0.63      &     34.36±0.24     &    \textbf{44.76±0.33}     &       38.87±0.21     &  44.96±2.69        &   55.25±2.77    &    49.57±2.73       \\
Struct              &      \textbf{53.89±0.78}        &      50.02±0.62     &      \textbf{51.88±0.69}    &      \textbf{62.12±0.41}      &     53.21±0.91     &     57.32±0.63   &       \textbf{47.07±0.15}       &    40.16±0.12      &       \textbf{43.34±0.10}   &     \textbf{57.61±1.87}    &  43.54±3.70   &     49.57±3.08    \\ \bottomrule
\end{tabular}}
\caption{Performance of state-of-art models on \textsc{Few-NERD (inter)}.}
\label{tab:inter}
\end{table*}

\subsection{The Overall Results}
We evaluate all baseline models on the three benchmark settings introduced in Section~\ref{sec:bench}, including \textsc{Few-NERD (sup)}, \textsc{Few-NERD (intra)} and \textsc{Few-NERD (inter)}.

\noindent \textbf{Supervised NER}\quad As mentioned in Section~\ref{sec:bench-supner}, we first split the $\textsc{Few-NERD}$ as a standard supervised NER dataset. As shown in Table~\ref{tab:sup}, BERT-Tagger yields promising results on the two widely used supervised datasets. The F1-score is 91.34\%, 89.11\%, respectively. However, the model suffers a grave drop in the performance on \textsc{Few-NERD (sup)} because the number of types of \textsc{Few-NERD (sup)} is larger than others. The results indicate that $\textsc{Few-NERD}$ is challenging in the supervised setting and worth studying.

We further analyze the performance of different entity types (see Figure~\ref{fig:types}). We find that the model achieves the best performance on the \texttt{Person} type and yields the worst performance on the \texttt{Product} type. And almost for all the coarse-grained types, the \texttt{Coarse-Other} type has the lowest F1-score. This is because the semantics of such fine-grained types are relatively sparse and difficult to be recognized. A natural intuition is that the performance of each entity type is related to the portion of the type. But surprisingly, we find that they are not linearly correlated. For examples, the model performs very well on the Art type, although this type represents only a small fraction of \textsc{Few-NERD}.

\noindent \textbf{Few-shot NER} \quad For the few-shot benchmarks, we adopt 4 sampling settings, which are $5$ way $1$$\sim$$2$ shot, $5$ way $5$$\sim$$10$ shot, $10$ way $1$$\sim$$2$ shot, and $10$ way $5$$\sim$$10$ shot. Intuitively, $10$ way $1$$\sim$$2$ shot is the hardest setting because it has the largest number of entity types and the fewest number of examples, and similarly, $5$ way $5$$\sim$$10$ shot is the easiest setting.
All results of $\textsc{Few-NERD (intra)}$ and $\textsc{Few-NERD (inter)}$ are reported in Table~\ref{tab:intra} and Table~\ref{tab:inter} respectively. Overall, we observe that the previous state-of-the-art methods equipped by BERT encoder could not yield promising results on $\textsc{Few-NERD}$. From a perspective of high level, models generally perform better on $\textsc{Few-NERD (inter)}$ than $\textsc{Few-NERD (intra)} $, and the latter is regarded as a more difficult task as we analyze in Section~\ref{sec:sim} and Section~\ref{sec:bench}, it splits the data according to the coarse-grained entity types, which means entity types between the training set and test set share less knowledge.

In a horizontal comparison, consistent with intuition, almost all the methods produce the worst results on $10$ way $1$$\sim$$2$ shot and achieve the best performance on $5$ way $5$$\sim$$10$. 
In the comparison across models, ProtoBERT generally achieves better performance than NNShot and StructShot, especially in $5$$\sim$$10$ shot setting where calculation by prototype may differ more from calculation by entity. StructShot has seen a large improvement in precision in $\textsc{Few-NERD (intra)}$. It shows that Viterbi decoder at the inference stage can help remove false positive predictions when knowledge transfer is hard. It is also observed that NNShot and StructShot may suffer from the instability of the nearest neighbor mechanism in the training phase, and prototypical models are more stable because the calculation of prototypes essentially serves as regularization.

\subsection{Error Analysis}
\vspace{-0.1cm}

We conduct error analysis to explore the challenges of $\textsc{Few-NERD}$, the results are reported in Table~\ref{tab:error}. We choose the setting of $\textsc{Few-NERD (inter)} $ because the test set contains all the coarse-grained types. We analyze the errors of models from two perspectives. \textit{Span Error} denotes the misclassifying in token-level classification. If an \texttt{O} token is misclassified as a part of entity, i.e., \texttt{I-type}, it is an FP case, and if a token with the type \texttt{I-type} is misclassified to \texttt{O}, it is FN. \textit{Type Error} indicates the misclassification of entity types when the spans are correctly classified. A ``Within'' error represents the entity is misclassified to another type within the same coarse-grained type, while ``Outer'' denotes the entity is misclassified to another type in a different coarse-grained type. As the statistics of type errors may be impacted by the sampled episodes in testing, we conduct 5 rounds of experiments and report the average results.
The results demonstrate that the token-level accuracy is not that low since most \texttt{O} tokens could be detected. But an entity mention is considered to be wrong if one token is wrong, which becomes the main reason for the challenge of $\textsc{Few-NERD}$. If an entity span could be accurately detected, the models could yield relatively good performance on entity typing, indicating the effectiveness of metric learning.

\begin{table}[]
\centering
\scalebox{0.88}{
\begin{tabular}{lllll} \toprule
\multirow{2}{*}{\textbf{Models}} & \multicolumn{2}{c}{\textbf{Span Error}}             & \multicolumn{2}{c}{\textbf{Type Error}}  \\ \cmidrule(r){2-3} \cmidrule(r){4-5}
                       & \multicolumn{1}{c}{\textbf{FP}} & \multicolumn{1}{c}{\textbf{FN}} & \multicolumn{1}{c}{\textbf{Within}} & \multicolumn{1}{c}{\textbf{Outer}} \\  \cmidrule(r){1-1} \cmidrule(r){2-3} \cmidrule(r){4-5}
ProtoNet               &            4.29\%            &            2.17\%            &             3.87\%               &            5.35\%               \\
NNShot                 &           3.87\%             &              3.67\%          &              3.86\%              &            6.90\%               \\
StructShot             &            2.84\%            &            4.45\%            &       3.94\%                     &    5.56\%     \\ \bottomrule                 
\end{tabular}}
\caption{Error analysis of 5 way 5$\sim$10 shot on $\textsc{Few-NERD (inter)}$, ``Within'' indicates ``within the coarse types'' and ``Outer'' is ``outer the coarse types''.}
\vspace{-0.2cm}
\label{tab:error}

\end{table}




\section{Conclusion and Future Work}

We propose $\textsc{Few-NERD}$, a large-scale few-shot NER dataset with fine-grained entity types. This is the first few-shot NER dataset and also one of the largest human-annotated NER dataset. $\textsc{Few-NERD}$ provides three unified benchmarks to assess approaches of few-shot NER and could facilitate future research in this area. 
By implementing state-of-the-art methods, we carry out 
a series of experiments on $\textsc{Few-NERD}$, demonstrating that few-shot NER remains a challenging problem and worth exploring. 
In the future, we will extend $\textsc{Few-NERD}$ by adding cross-domain annotations, distant annotations, and finer-grained entity types. \textsc{Few-NERD} also has the potential to advance the construction of continual knowledge graphs.

\section*{Acknowledgements}

This research is supported by  National Natural Science Foundation of China (Grant No. 61773229 and 6201101015),  National Key Research and Development Program of China (No. 2020AAA0106501), Alibaba Innovation Research (AIR) programme, the General Research Project (Grand No. JCYJ20190813165003837, No.JCYJ20190808182805919), and Overseas Cooperation Research Fund of Graduate School at Tsinghua University (Grant No. HW2018002). Finally, we thank the valuable help of Ronny, Xiaozhi, Ziyu and comments of anonymous reviewers.



\section*{Ethical Considerations}
In this paper, we present a human-annotated dataset, $\textsc{Few-NERD}$, for few-shot learning in NER. We describe the details of the collection process and conditions, the compensation of annotators, the measurements to ensure the quality in the main text. The corpus of the dataset is publicly obtained from Wikipedia and we have not modified or interfered with the content.
$\textsc{Few-NERD}$ is likely to directly facilitate the research of few-shot NER, and further increase the progress of the construction of large-scale knowledge graphs (KGs). Models and systems built on $\textsc{Few-NERD}$ may contribute to construct  KGs in various domains, including biomedical, financial, and legal fields, and further promote the development of NLP applications on specific domains. 
$\textsc{Few-NERD}$ is annotated in  English, thus the dataset may mainly facilitate NLP research in English. For the sake of energy saving, we will not only open source the dataset and the code, but also release the checkpoints of our models from the experiments to reduce unnecessary carbon emission.

\bibliographystyle{acl_natbib}
\bibliography{anthology,acl2021}

\clearpage
\appendix
\section{Data Details}

\subsection{Processing}
We use the dump\footnote{\url{https://dumps.wikimedia.org/enwiki/}} of English Wikipedia, and extract the raw text by WikiExtractor\footnote{\url{https://github.com/attardi/wikiextractor}}. NLTK language tool\footnote{\url{https://www.nltk.org}} is used for word and sentence tokenization in the preprocessing stage. As stated in Section~\ref{sec:para}, we develope a search engine to index and select paragraphs with key words in distant dictionaries. If the search is performed with linear operations, the calculation process will be extremely slow, instead, we adopt a search engine with Lucene\footnote{\url{https://lucene.apache.org/}} to conduct effective indexing and searching.

\subsection{More Details of the Schema}
As stated in Section 4.1,  we use $\textsc{Figer}$~\cite{ling2012fine} as the start point and conduct rounds of  make a series of modifications. Despite the modifications mentioned in Section~\ref{sec:schema}, we also conduct manual denoising of the automatically annotated data of $\textsc{Fier}$. For each entity type and the corresponding automatically annotated mentions, we randomly select 500 mentions and compute the accuracy to obtain the real frequency. For example, statistics report that \texttt{cemetery} is a type with high frequency. However, a plenty number of the mentions labeled as \texttt{cemetery} are actually \texttt{GPE}. Similarly, \texttt{engineer} is also affected by noise.

\subsection{Interface}
\begin{figure}[h]
    \centering
    \includegraphics[width = 0.95\linewidth]{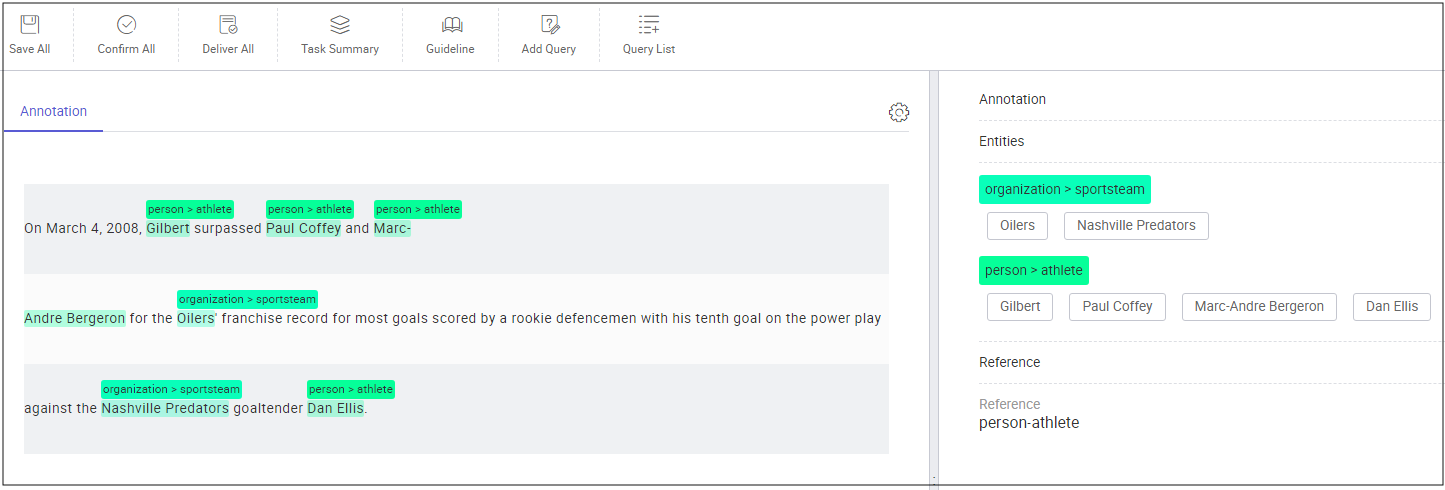}
    \caption{Screeshot of the interface used to annotate \textsc{Few-NERD}.}
    \label{fig:inter}
\end{figure}

The interface in shown in Figure~\ref{fig:inter}, where annotators could expediently select entity spans and annotate the corresponding coarse and fine types. And annotators could check the current annotation information on the interface.

\section{Implementation Details}

All the four models use $\text{BERT}_{\text{base}}$~\cite{devlin2018bert} and the backbone encoder and initialized with the corresponding pre-trained uncased weights\footnote{\url{https://github.com/google-research/bert}}. 
The hidden size is 768, and the number of layers and heads are 12.  Models are implemented by Pytorch framework\footnote{\url{https://pytorch.org}}~\cite{paszke2019pytorch} and  Huggingface transformers\footnote{\url{https://github.com/huggingface/transformers}}~\cite{wolf2019huggingface}.
BERT models are optimized by AdamW\footnote{\url{https://www.fast.ai/2018/07/02/adam-weight-decay/\#adamw}}~\cite{loshchilov2018decoupled} with the learning rate of 1e-4. We evaluate our implementations of NNShot and StructShot on the datasets used in the original paper, producing similar results.
For supervised NER, the batch size is 8, and we train BERT-Tagger for 70000 steps and evaluate it on the test set.
For 5 way 1$\sim$2 and 5$\sim$10 shot settings, the batch sizes are 16 and 4, and for 10 way 1$\sim$2 and 5$\sim$10 shot settings, the batch sizes are 8 and 1. We train 12000 episodes and use 500 episodes of the dev set to select the best model, and test it on 5000 episodes of the test set. Most hyper-parameters are from original settings. We manually tune the hyper-parameter \textit{$\tau$} in Viterbi for StructShot, and the value for 1$\sim$2 settings shot is 0.320, for 5$\sim$10 shot settings is 0.434.
All the experiments are conducted with CUDA on NVIDIA Tesla V100 GPUs. With 2 GPUs used, the average time to train 10000 episodes is 135 minutes. The number of parameters of the models is 120M.

\section{Entity Types}
As introduced in Section~\ref{sec:schema} in main text, $\textsc{Few-NERD}$ is manually annotated with 8 coarse-grained and 66 fine-grained entity types, and we list all the types in Table 8. The schema is designed under practical situation, we hope the schema could help to better understand $\textsc{Few-NERD}$. Note that \texttt{ORG} is the abbreviation of \texttt{Organization}, and \texttt{MISC} is the abbreviation of \texttt{Miscellaneous}.

\onecolumn

\footnotesize
\newpage
\begin{longtable}{clp{9.5cm}} 

\toprule
\textbf{Coarse Type} & \textbf{Fine Type} & \textbf{Example} \\ \midrule

\multicolumn{1}{l}{\multirow{7}{*}{\texttt{\color{BlueViolet}Location}}} & \texttt{\color{BlueViolet}GPE}                 & The company moved to a new office in \textit{\color{BlueViolet}Las Vegas}, \textit{Nevada}.                                                                                  \\ \cmidrule(r){2-2} \cmidrule(r){3-3}
\multicolumn{1}{c}{}                          & \texttt{\color{BlueViolet}Body of Water}     & The \textit{\color{BlueViolet}Finke River} normally drains into the Simpson Desert to the north west of the Macumba.                                                      \\ \cmidrule(r){2-2} \cmidrule(r){3-3}
\multicolumn{1}{c}{}                          & \texttt{\color{BlueViolet}Island}              & An invading army of Teutonic Knights conquered \textit{\color{BlueViolet}Gotland} in 1398.                                                                                \\ \cmidrule(r){2-2} \cmidrule(r){3-3}
\multicolumn{1}{c}{}                          & \texttt{\color{BlueViolet}Mountain}            & C.G.E. Mannerheim met Thubten Gyatso in \textit{\color{BlueViolet}Wutai Shan} during the course of his expedition from Turkestan to Peking.                               \\ \cmidrule(r){2-2} \cmidrule(r){3-3}
\multicolumn{1}{c}{}                          & \texttt{\color{BlueViolet}Park}                & \textit{\color{BlueViolet}Victoria Park} contains examples of work by several architects including Alfred Waterhouse (Xaverian College).                                  \\ \cmidrule(r){2-2} \cmidrule(r){3-3}
\multicolumn{1}{c}{}                          & \texttt{\color{BlueViolet}Road/Transit}       & The thirty-first race of the 1951 season was held on October 7 at the one-mile dirt \textit{\color{BlueViolet}Occoneechee Speedway}.                                      \\ \cmidrule(r){2-2} \cmidrule(r){3-3}
\multicolumn{1}{c}{}                          & \texttt{\color{BlueViolet}Other}               & Herodotus (7.59) reports that \textit{\color{BlueViolet}Doriscus} was the first place Xerxes the Great stopped to review his troops.                                      \\ \midrule
\multirow{8}{*}{\texttt{\color{PineGreen}Person}}                       & \texttt{\color{PineGreen}Actor}               & The first performance of any work of \textit{\color{PineGreen}Gustav Holst} given in that capital.                                                                       \\ \cmidrule(r){2-2} \cmidrule(r){3-3}
                                              & \texttt{\color{PineGreen}Artist/Author}      & A film adaption was made by \textit{\color{PineGreen}Arne Bornebusch} in 1936.                                                                                           \\ \cmidrule(r){2-2} \cmidrule(r){3-3}
                                              & \texttt{\color{PineGreen}Athlete}             & \textit{\color{PineGreen}Smith} was named co-Player of the Week in the Big Ten on offense.                                                                               \\ \cmidrule(r){2-2} \cmidrule(r){3-3}
                                              & \texttt{\color{PineGreen}Director}            & Margin for Error is a 1943 American drama film directed by \textit{\color{PineGreen}Otto Preminger}.                                                                     \\ \cmidrule(r){2-2} \cmidrule(r){3-3}
                                              & \texttt{\color{PineGreen}Politician}          & Then-President \textit{\color{PineGreen}Gloria Macapagal Arroyo} led the inauguration rites of the facility on August 19, 2002.                                          \\ \cmidrule(r){2-2} \cmidrule(r){3-3}
                                              & \texttt{\color{PineGreen}Scholar}             & \textit{\color{PineGreen}Jeffery Westbrook} and \textit{\color{PineGreen}Robert Tarjan} (1992) developed an efficient data structure for this problem based on disjoint-set data structures. \\ \cmidrule(r){2-2} \cmidrule(r){3-3}
                                              & \texttt{\color{PineGreen}Soldier}             & \textit{\color{PineGreen}Sadowski} was promoted to general, and took command of the freshly created Fortified Area of Silesia.                                           \\ \cmidrule(r){2-2} \cmidrule(r){3-3}
                                              & \texttt{\color{PineGreen}Other}               & In Albany, \textit{\color{PineGreen}Doane} planned a cathedral like those in England.                                                                                    \\ \midrule
\multirow{10}{*}{\texttt{\color{Mahogany}ORG}}                & \texttt{\color{Mahogany}Company}             & A Vocaloid voicebank developed and distributed by \textit{\color{Mahogany}Yamaha Corporation} for Vocaloid 4.                                                           \\ \cmidrule(r){2-2} \cmidrule(r){3-3} 
                                              & \texttt{\color{Mahogany}Education}           & Long volunteer coached the offensive line for \textit{\color{Mahogany}Briarcrest Christian School} for 9 seasons.                                                       \\ \cmidrule(r){2-2} \cmidrule(r){3-3}
                                              & \texttt{\color{Mahogany}Government}          & It was constructed using the savings of the \textit{\color{Mahogany}Quezon provincial government}.                                                                      \\ \cmidrule(r){2-2} \cmidrule(r){3-3}
                                              & \texttt{\color{Mahogany}Media}               & He was the Editor in Chief of Grenada’s national newspaper "\textit{\color{Mahogany}The Free West Indian}".                                                             \\ \cmidrule(r){2-2} \cmidrule(r){3-3}
                                              & \texttt{\color{Mahogany}Political/party}    & Stanley Norman Evans was a British industrialist and \textit{\color{Mahogany}Labour Party} politician.                                                                  \\ \cmidrule(r){2-2} \cmidrule(r){3-3}
                                              & \texttt{\color{Mahogany}Religion}            & D'Souza was born on 10 November 1985 into a \textit{\color{Mahogany}Goan Catholic} family in Goa, India.                                                                \\ \cmidrule(r){2-2} \cmidrule(r){3-3}
                                              & \texttt{\color{Mahogany}Sports League}      & His strong performances convinced him that he was ready for the \textit{\color{Mahogany}NBA}.                                                                           \\ \cmidrule(r){2-2} \cmidrule(r){3-3}
                                              & \texttt{\color{Mahogany}Sports Team}        & \textit{\color{Mahogany}The Pirates} won the game and the World Series with Oldham on the mound.                                                                        \\ \cmidrule(r){2-2} \cmidrule(r){3-3}
                                              & \texttt{\color{Mahogany}Show ORG}  & Standing in the Way of Control is the third studio album by American indie rock band \textit{\color{Mahogany}Gossip}.                                                   \\ \cmidrule(r){2-2} \cmidrule(r){3-3}
                                              & \texttt{\color{Mahogany}Other}               & He is the Creative Director of the \textit{\color{Mahogany}Oliver Sacks Foundation}.                                                                                    \\ \midrule
\multirow{8}{*}{\color{TealBlue}\texttt{Building}}                     & \texttt{\color{TealBlue}Airport}             & The city is served by the \textit{\color{TealBlue}Sir Seretse Khama International Airport}.                                                                             \\ \cmidrule(r){2-2} \cmidrule(r){3-3}
                                              & \texttt{\color{TealBlue}Hospital}            & Then he did residency in ophthalmology at \textit{\color{TealBlue}Farabi Eye Hospital} from 1979 to 1982.                                                               \\ \cmidrule(r){2-2} \cmidrule(r){3-3}
                                              & \texttt{\color{TealBlue}Hotel}               & Nick also played at the regular Sunday evening sessions that were held at the \textit{\color{TealBlue}Ramada Inn} in Schenectady.                                       \\ \cmidrule(r){2-2} \cmidrule(r){3-3}
                                              & \texttt{\color{TealBlue}Library}             & \textit{\color{TealBlue}RMIT University Library} consists of six academic branch libraries in Australia and Vietnam.                                                    \\ \cmidrule(r){2-2} \cmidrule(r){3-3}
                                              & \texttt{\color{TealBlue}Restaurant}          & The first \textit{\color{TealBlue}Panda Express restaurant} opened in Galleria II in the same year, on level 3 near Bloomingdale's.                                     \\ \cmidrule(r){2-2} \cmidrule(r){3-3}
                                              & \texttt{\color{TealBlue}Sports Facility}    & This was the last year that the Razorbacks would play in \textit{\color{TealBlue}Barnhill Arena}.                                                                       \\ \cmidrule(r){2-2} \cmidrule(r){3-3}
                                              & \texttt{\color{TealBlue}Theater}             & From 1954, she became a guest singer at the \textit{\color{TealBlue}Vienna State Opera}.                                                                                \\ \cmidrule(r){2-2} \cmidrule(r){3-3}
                                              & \texttt{\color{TealBlue}Other}               & Eissler designated Masson to succeed him as Director of the \textit{\color{TealBlue}Sigmund Freud Archives} after his and Anna Freud's death.                           \\ \midrule
\multirow{6}{*}{\texttt{\color{Plum}Art}}                          & \texttt{\color{Plum}Music}               & "\textit{\color{Plum}Get Right}" is a song recorded by American singer Jennifer Lopez for her fourth studio album.                                                  \\ \cmidrule(r){2-2} \cmidrule(r){3-3}
                                              & \texttt{\color{Plum}Film}                & \textit{\color{Plum}Margin for Error} is a 1943 American drama film directed by Otto Preminger.                                                                     \\ \cmidrule(r){2-2} \cmidrule(r){3-3}
                                              & \texttt{\color{Plum}Written Art}        & \textit{\color{Plum}The Count} is a text adventure written by Scott Adams and published by Adventure International in 1979.                                         \\ \cmidrule(r){2-2} \cmidrule(r){3-3}
                                              & \texttt{\color{Plum}Broadcast}           & In the fall of 1957, Mitchell starred in ABC's "\textit{\color{Plum}The Guy Mitchell Show}".                                                                        \\ \cmidrule(r){2-2} \cmidrule(r){3-3}
                                              & \texttt{\color{Plum}Painting}            & His painting '\textit{\color{Plum}Rooftops}' has been in the collection of the City of London Corporation since 1989.                                               \\ \cmidrule(r){2-2} \cmidrule(r){3-3}
                                              & \texttt{\color{Plum}Other}               & Kirwan appeared on stage at the Chichester Festival Theatre in a Jeremy Herrin production of \textit{\color{Plum}Uncle Vanya}.                                      \\ \midrule
\multirow{9}{*}{\texttt{\color{Red}Product}}                      & \texttt{\color{Red}Airplane}            & The Royal Norwegian Air Force's 330 Squadron operates a \textit{\color{Red}Westland Sea King} search and rescue helicopter out of Florø.                           \\ \cmidrule(r){2-2} \cmidrule(r){3-3}
                                              & \texttt{\color{Red}Car}                 & The BYD \textit{\color{Red}Tang} plug-in hybrid SUV was the top selling plug-in car with 31,405 units delivered.                                                   \\ \cmidrule(r){2-2} \cmidrule(r){3-3}
                                              & \texttt{\color{Red}Food}             & The words "Time to make the donuts" are printed on the side of \textit{\color{Red}Dunkin' Donuts} boxes in memory of Michael Vale/Fred the Baker.                  \\ \cmidrule(r){2-2} \cmidrule(r){3-3}
                                              & \texttt{\color{Red}Game}                & Team Andromeda wanted to create a fully 3D arcade game, having worked on similar games such as "\textit{\color{Red}Out Run}" which were not truly 3D.              \\ \cmidrule(r){2-2} \cmidrule(r){3-3}
                                              & \texttt{\color{Red}Ship}                & As night fell, Marine Corps General Holland Smith studied reports aboard the command ship "\textit{\color{Red}Eldorado}".                                          \\ \cmidrule(r){2-2} \cmidrule(r){3-3}
                                              & \texttt{\color{Red}Software}            & It allows communication between the \textit{\color{Red}Wolfram Mathematica} kernel and front-end.                                                                  \\ \cmidrule(r){2-2} \cmidrule(r){3-3}
                                              & \texttt{\color{Red}Train}               & On 9 June 1929, railcar No. 220 "\textit{\color{Red}Waterwitch}" overran signals at Marshgate Junction.                                                            \\ \cmidrule(r){2-2} \cmidrule(r){3-3}
                                              & \texttt{\color{Red}Weapon}              & Mannerheim gave Tibet's spiritual pontiff a \textit{\color{Red}Browning revolver} and showed him how to reload the weapon.                                         \\ \cmidrule(r){2-2} \cmidrule(r){3-3}
                                              & \texttt{\color{Red}Other}               & \textit{\color{Red}Rhinestone} is as artificial and synthetic a concoction as has ever made its way to the screen.                                                 \\ \midrule
\multirow{6}{*}{\texttt{\color{Peach}Event}}                        & \texttt{\color{Peach}Attack}              & It was on this route that Tecumseh was killed at the \textit{\color{Peach}Battle of the Thames} on October 5, 1813.                                                  \\ \cmidrule(r){2-2} \cmidrule(r){3-3}
                                              & \texttt{\color{Peach}Election}            & At the \textit{\color{Peach}1935 United Kingdom general election}, McGleenan stood in Armagh as an Independent Republican.                                           \\ \cmidrule(r){2-2} \cmidrule(r){3-3}
                                              & \texttt{\color{Peach}Natural Disaster}   & He was originally from Chicago, but moved to Japan after the \textit{\color{Peach}Second Great Kanto earthquake} that all but decimated Japan's infrastructure.      \\ \cmidrule(r){2-2} \cmidrule(r){3-3}
                                              & \texttt{\color{Peach}Protest}             & In 1832, following the failed \textit{\color{Peach}Polish November Uprising}, the Dominican monastery was sequestrated.                                              \\ \cmidrule(r){2-2} \cmidrule(r){3-3}
                                              & \texttt{\color{Peach}Sports Event}       & Carle received a new defense partner when the Flyers traded for Chris Pronger at the 2009 \textit{\color{Peach}NHL Entry Draft}.                                     \\ \cmidrule(r){2-2} \cmidrule(r){3-3}
                                              & \texttt{\color{Peach}Other}               & One of TMG's first performances was in September 1972 at the \textit{\color{Peach}Waitara Festival}.                                                                 \\ \midrule
\multirow{12}{*}{\color{Brown}\texttt{MISC}}                       & \texttt{\color{Brown}Astronomy}           & He discovered a number of double stars and took many photographs of \textit{\color{Brown}Mars}.                                                                      \\ \cmidrule(r){2-2} \cmidrule(r){3-3}
                                              & \texttt{\color{Brown}Award}               & He was awarded \textit{\color{Brown}the Bialik Prize} eight years later for these efforts.                                                                           \\ \cmidrule(r){2-2} \cmidrule(r){3-3}
                                              & \texttt{\color{Brown}Biology}             & \textit{\color{Brown}Estradiol valerate} is rapidly hydrolyzed into \textit{\color{Brown}estradiol} in the intestines.                                                               \\ \cmidrule(r){2-2} \cmidrule(r){3-3}
                                              & \texttt{\color{Brown}Chemistry}           & It was the first gas manufacturer in Kuwait to provide industrial gases such as \textit{\color{Brown}\color{Brown}oxygen} and \textit{\color{Brown}nitrogen} to the local petroleum industry.     \\ \cmidrule(r){2-2} \cmidrule(r){3-3}
                                              & \texttt{\color{Brown}Currency}            & Total investment has been 19 billion \textit{\color{Brown}Norwegian krone}.                                                                                          \\ \cmidrule(r){2-2} \cmidrule(r){3-3}
                                              & \texttt{\color{Brown}Disease}             & The 2020 competition was cancelled as part of the effort to minimize the \textit{\color{Brown}COVID-19} pandemic.                                                    \\ \cmidrule(r){2-2} \cmidrule(r){3-3}
                                              & \texttt{\color{Brown}Educational Degree} & Sigurlaug enrolled into the medical department of the University of Iceland and graduated as a \textit{\color{Brown}Medical Doctor} in 2010.                         \\ \cmidrule(r){2-2} \cmidrule(r){3-3}
                                              & \texttt{\color{Brown}God}                 & Originally a farmer, Viking Ragnar Lothbrok claims to be descended from the god \textit{\color{Brown}Odin}.                                                          \\ \cmidrule(r){2-2} \cmidrule(r){3-3}
                                              & \texttt{\color{Brown}Language}            & The play was translated into \textit{\color{Brown}English} by Michael Hofmann and published in 1987 by Hamish Hamilton.                                              \\ \cmidrule(r){2-2} \cmidrule(r){3-3}
                                              & \texttt{\color{Brown}Law}                 & Four of his five policy recommendations were incorporated into \textit{\color{Brown}the U.S. Federal Financial Law} of 1966.                                         \\ \cmidrule(r){2-2} \cmidrule(r){3-3}
                                              & \texttt{\color{Brown}Living Thing}       & \textit{\color{Brown}Schistura horai} is a species of ray-finned fish in the stone loach genus "\textit{\color{Brown}Schistura}".                                                    \\ \cmidrule(r){2-2} \cmidrule(r){3-3}
                                              & \texttt{\color{Brown}Medical}             & Precious Blood Hospital offers specialist outpatient and inpatient services in \textit{\color{Brown}General medicine}.   \\ \bottomrule       
                                              
\caption{All the coarse-grained and fine-grained entity types in $\textsc{Few-NERD}$, we only highlight the entities with the corresponding entity types in ``Example''.}
\end{longtable}

\end{document}